\pdfoutput=1

\documentclass[11pt]{article}

\usepackage[final]{acl}
\usepackage{microtype}
\usepackage{hyperref}
\usepackage{url}
\usepackage{booktabs}
\usepackage{microtype}
\usepackage{hyperref}
\usepackage{url}
\usepackage{booktabs}
\usepackage{booktabs}
\usepackage{subcaption}
\usepackage{algorithm}
\usepackage{times}
\usepackage{textcomp}
\usepackage{xcolor}
\usepackage{amsmath,amssymb,amsfonts}
\usepackage{graphicx}
\usepackage{algpseudocode}
\usepackage{float}
\usepackage{listings}
\usepackage[export]{adjustbox}
\usepackage{endnotes}

\title{Semantic Convergence: Investigating Shared Representations Across Scaled LLMs}

\newcommand{\authornote}[1]{\endnote{#1}}

\author{
  Daniel Son\authornote{\url{gtb4ua@virginia.edu}}
  \And
  Sanjana Rathore\authornote{\url{sanjanar@g.ucla.edu}}
  \And
  Andrew Rufail\authornote{\url{andrew.rufail@gmail.com}}
  \AND
  Adrian Simon\authornote{\url{adriansimon477@gmail.com}}
  \And
  Daniel Zhang\authornote{\url{dtz2104@columbia.edu}}
  \And
  Soham Dave\authornote{\url{davesoham14@gmail.com}}
  \AND
  Cole Blondin\authornote{\url{cole@algoverseairesearch.org}}
  \And
  Kevin Zhu\authornote{\url{kevin@algoverseacademy.com}}
  \And
  Sean O’Brien\authornote{\url{seobrien@ucsd.edu}}
}

\begin{document}
\maketitle

\begin{abstract}
We investigate feature universality in Gemma-2 language models (Gemma-2-2B \& Gemma-2-9B), asking whether models with a fourfold difference in scale still converge on comparable internal concepts. Using the sparse autoencoder (SAE) dictionary learning pipeline, we used pretrained SAEs on each model’s residual-stream activations, aligned the resulting monosemantic features via activation correlation, and compared the matched feature spaces with metrics such as SVCCA and RSA. Middle layers yield the strongest overlap, indicating that this is where both models most similarly represent concepts, while early and late layers show much less similarity. Preliminary experiments extending the analysis from single tokens to multi-token subspaces show that semantically similar subspaces tend to interact similarly with LLMs. These results offer further evidence that large language models carve the world into broadly similar, interpretable features despite size differences, reinforcing universality as a foundation for cross-model interpretability.

\end{abstract}

\section{Introduction}
Large Language Models (LLMs) \citep{achiam2023gpt, touvron2023llama, guo2025deepseek} have demonstrated increasing reasoning abilities across many tasks \citep{bubeck2023sparksartificialgeneralintelligence}. However, our understanding of the internal representations and computations that support this behavior remains limited \citep{bereska2024mechanisticinterpretabilityaisafety}.\\
\\Previous work \citep{lan2024sparse} has shown that models with the same tokenizer rely on similar internal representations and structures, indicating that universal feature spaces might exist. We define a feature as being universal if its activation corresponds to the same semantic concept regardless of model size or architecture. Universal feature spaces may be encoded across different layers or neurons in different models, but there would exist a sparse direction in each model’s activation space that consistently “lights up” on these concepts. Understanding universal feature spaces is a crucial step in learning if general rules govern how LLMs internally structure and use their nodes. This key would increase the generalizability of interpreting different LLMs and may accelerate LLM training efficiency as well as LLM safety \citep{chughtai2023toymodeluniversalityreverse, gurnee2024universalneuronsgpt2language, bricken2023monosemanticity}.\\
 \\
Comparing features between LLMs is challenging because nodes in the model usually represent multiple features, rather than one specific feature. This is called polysemanticity \citep{elhage2022superposition}. In this paper, we build on the methods shown in \citet{lan2024sparse} that leverage Sparse Autoencoders (SAEs) to transform LLM node activations into lower dimensional spaces that are easier to interpret. The main advantage of using SAEs is that they have the ability to decompose the complex, polysemantic representations in an LLM into distinct features that can be interpreted more easily \citep{cunningham2023sparseautoencodershighlyinterpretable, bricken2023monosemanticity}. Then, representational space similarity metrics are used on these SAE features to check for similarities in the internal structure of the LLM. \\

Although results for feature universality in \citet{lan2024sparse} were promising, only single token words, in a limited number of semantic subspaces, were tested for the semantic experiments. Furthermore, the experiments were only carried out against similar sized models, namely Pythia-70m with Pythia-160m \citep{biderman2023pythiasuiteanalyzinglarge} and Gemma-1-2B with Gemma-2-2B \citep{gemmateam2024gemma2improvingopen, gemmateam2024gemmaopenmodelsbased}. Therefore, in this paper, we will further investigate the universality of feature spaces. \\

In our experiments, we use models with a four fold size difference. Our results demonstrate that the similarity in internal feature representations remains across these models despite difference in complexity. Furthermore, in our semantic subspaces studies, we show that there are certain groups of overlapping concepts that the models internally represent similarly. This is another indication of feature universality. \\
In summary, our contributions are the following:
\begin{enumerate}
    \item Probe universality in multi-token semantic subspaces, including overlaps of related concepts, to see whether phrase-level and higher-order features align across models.
    \item Quantify universality across a 4× model-size gap and compare similarity measures (e.g., SVCCA, RSA) to test how metric choice affects the result.
\end{enumerate}
Our results reveal that, despite the fourfold size difference, models share strongly aligned internal representations. In addition, we demonstrate that conceptually coherent semantic groups align. We also illustrate that models encode multi-token phrases across the layers, where phrases are represented more strongly than individual tokens in the earlier layers.\\

Understanding whether different-sized language models develop similar internal representations has significant practical implications. If models trained independently converge to shared semantic features, this suggests the existence of a common representational structure, opening the door to transferable interpretability tools, cross-scale safety interventions \citep{engels2025scaling}, and more efficient model distillation pipelines \citep{turc2019well}. These insights could fundamentally reshape how we evaluate, interpret and design future language models.\\

This result supports the idea that universality is a structural property of the Gemma family, not an accident of size or training noise. It gives practitioners a clear path to build and test interpretability tools on the lighter 2 B model before deploying them unchanged on the 9 B model. Similar mid-layer convergence has also been reported for the Pythia family (70 M → 160 M) by \citet{lan2024sparse}.  Taken together with our Gemma results, this suggests that feature universality extends across at least two independent model families.\\

This work opens up several branches of future research that we believe are worth studying. Training SAEs on multiple model layers can reveal internal representations that are not captured in a single layer. In addition, comparing the internal representations of SAEs trained on MLP layers may provide deeper insights about the universality of MLP features. These findings can accelerate AI reasoning and safety training \citep{hendrycks2023overviewcatastrophicairisks}. Through understanding the similarities between models, and their differences, a more complete picture of how LLMs process, reason and understand natural language would be formed \citep{lan2024sparse}. 

\begin{figure*}
  \centering
  \includegraphics[scale=0.6]{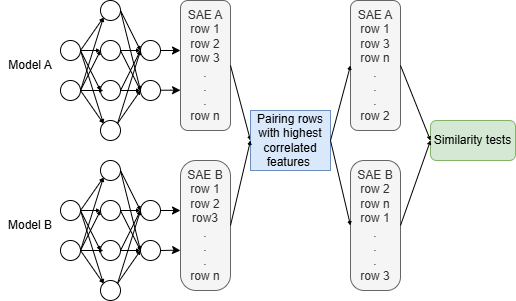} 
  \caption{Workflow of pairing rows with the highest correlated features between two models (Gemma-2-2B and Gemma-2-9B) and performing similarity tests to assess feature alignment.}
  \label{fig:example}
\end{figure*}
\section{Background}
\textbf{Sparse Autoencoders}. Sparse Autoencoders (SAEs) are a type of neural network used to learn efficient, sparse representations of input data \citep{Makhzani2013kSparseA}. Unlike other autoencoders, SAEs incorporate a sparsity constraint, typically an L1 penalty on the hidden layer activations or a KL divergence term, which pushes most hidden units to be inactive (i.e. any output values close to zero) for any input given. This leads to features that are more interpretable and disentangled. The aim is to discover a basis of features, similar to dictionary learning \citep{OLSHAUSEN19973311}, where each feature activates for semantically meaningful concepts.
\\Mathematically, an input $x \in \mathbb{R}^n$ is given to the neural network which is reconstructed into $\hat{x}$ using $\hat{x} = W'\sigma(Wx + b)$, where $W \in \mathbb{R}^{hxn}$ is the encoder weight matrix, b is the bias term, $\sigma$ is a nonlinear activation function, and $W'$ is the decoder matrix, which often uses the transpose of the encoder weights. SAE training seeks to both encourage sparsity in the activations $h = \sigma(Wx + b)$ and to minimize the reconstruction loss $L_{\text{rec}}(x, \hat{x}) = \| x - \hat{x} \|^2$.\\

\section{Methods}
\subsection{Feature Pairings}
To determine whether different models of varying sizes converge on similar internal representations, generalizations of feature spaces, spaces formed by feature groups, and feature relations must be explored. To quantitatively measure these similarities, we follow the methods of \citet{lan2024sparse}. Overall, we compare an SAE trained on layer $A_i$ from LLM A with another SAE trained on layer $B_j$ from LLM B for every layer pairing.\\
\\
However, accurate comparisons between spaces hinges on solving two issues:\\
\textbf{Permutation issue}. To solve the permutation issue, we  find neuron pairings that are the most similar in $SAE_A$ and $SAE_B$. Since the mapping of features is unknown due to arbitrary neuron indexing and some features may not have a ``similar'' feature in the other SAE, we pairwise match them using a correlation metric. \\ 
\textbf{Rotational Alignment issue}. Even after permutation alignment, each SAE may use its own orthonormal basis for latent space. To ensure that the true relational similarity is captured, we apply rotation‑invariant similarity measures, namely SVCCA and RSA. \\ 
\\
To score the results against a baseline, randomly paired features are obtained. Then, the score of the features paired by correlation
(referred to as ``paired features'') is compared with the average score of $N$ runs of randomly paired features to obtain a p-value score.
\subsubsection{1-to-1 vs.\ Many-to-1 Feature Matching}
\label{sec:matching}

Following \citet{lan2024sparse}, we consider two ways of pairing SAE
features from layer~$A_i$ of the first model with layer~$B_j$ of the
second.
\paragraph{1-to-1 (bijective) matching.}
We iteratively build a one–to–one assignment: at each step we pick the
still–unmatched pair of features with the highest Pearson
correlation.  Each feature is used at most once, yielding a bijection of
size $K=\min\!\bigl(|A_i|,|B_j|\bigr)$.
\paragraph{Many-to-1 matching.}
To probe whether the entire dictionary of the smaller layer can be
embedded inside the larger one, we relax the uniqueness constraint on
layer~$B_j$.  Every feature in $A_i$ is matched to its most-correlated
partner in $B_j$, even if that target has already been claimed by other
sources.  Thus one feature in $B_j$ may receive multiple links, while
each feature in $A_i$ is still matched exactly once.

Unless stated otherwise, all correlations are computed with Pearson
correlation; the aligned pairs returned by the chosen strategy are then
fed into the subsequent SVCCA and RSA calculations.

\subsection{Representational Similarity Metrics}
\subsubsection{Singular Value Canonical Correlation Analysis (SVCCA)} Singular Value Canonical Correlation Analysis \citep{Raghu2017SVCCASV} is a variation of the Canonical Correlation Analysis CCA \citep{hotelling1936relations} which finds a pair of the most correlated variables, $u_i$ and $v_i$ from two sets of variables $X \in \mathbb{R}^{n \times d_1}$ and $Y \in \mathbb{R}^{n \times d_2}$. Before applying CCA, SVCCA reduces noise by applying Singular Value Decomposition (SVD) to $X$ and $Y$ using $X=U_XS_XV_X^T$ and $Y=U_YS_YV_Y^T$, where $U_X$ and $U_Y$ are the matrices containing the left singular vectors (informative directions), and $S_X$ and $S_Y$ are diagonal matrices containing the singular values. After CCA is applied on the new data, correlation scores between the most informative components are obtained, which are then averaged to get a similarity score. SVCCA measures how well subspaces of two SAE weight matrices align, essentially quantifying the global feature space overlap.

\subsubsection{Representational Similarity Analysis (RSA)} Representational Similarity Analysis \citep{Kriegeskorte2008} calculates, for each space, a Representation Dissimilarity Matrix (RDM) $D \in \mathbb{R}^{nxn}$. Each element in this matrix represents the dissimilarity between every pair of data points in the space. Following RDM, a correlation metric such as Spearman's rank correlation coefficient is used to compute a similarity score.

\subsection{Semantic Subspaces}
In addition to the layer-wise SAE comparisons, we also test semantic subspaces, collections of words defined by a high-level concept that contain concept-specific keywords. For example, ``emotions'' is a subspace with concept-specific keywords like ``happy''and ``sad''. By testing these subspaces, we can evaluate whether LLMs encode the same semantic categories.\\

For each high-level concept, we first use GPT‑4o \citep{achiam2023gpt} to generate three independent lists of representative keywords. We then intersect these lists and retain only terms that are unambiguous (each having a single and clear meaning). Next, we add to the keyword set their hyponyms from WordNet \citep{wordnet}. This combined collection of keywords plus and their hyponyms defines the final semantic subspace for that concept.\\

To evaluate semantic subspaces more rigorously, we combine two different subspaces in the following ways:\\
\textbf{Multi-token subspaces:} In the multi-token subspaces, we concatenate keywords from different concepts together. For instance, ``happy'' (from ``emotions'') and ``child'' (from ``person'') becomes ``happy child''. In these types of subspaces, we aim to understand if different LLMs internally process longer sentences similarly. Furthermore, we concatenated unlikely pairs of concepts, such as ``calendar'' and ``emotions'', which the LLMs were unlikely to see during their training to check whether the LLMs process previously unseen data in a similar manner.\\
\textbf{Overlapping subspaces:} Overlapping subspaces are formed by taking the union of whole subspaces together. For instance, the ``emotions'' and ``person'' subspaces would yield (``happy'', ``teacher'', ``sad'', ``child'', ...). This aims to test if the different LLMs interact with multiple concepts similarly.\\

\section{Experiments and Results}
\subsection{Feature Analysis}
\textbf{Layer-wise similarity of full SAE spaces}

\textbf{1-to-1 (5 run mean):} In Fig. 2a the diagonal band of Paired SVCCA now peaks at 0.73 (Gemma-2-2B L14 and Gemma-2-9B L19) and stays consistently high across contiguous mid-layer pairs (0.64 – 0.71). Early layers sit at 0.35 ± 0.05, and the last decoder layer of Gemma-2-9B (L39) experiences a drop to an average of 0.374. Through these visualizations (Fig. 2a \& Fig. 2b), it is observable that the middle layers between both models share the most similarity compared to other layers.

Paired RSA (Fig. 2b) follows the same shape but at roughly one-third of the magnitude: maxima of 0.22 and a midlayer plateau of 0.15 – 0.20, with edges staying less than 0.08. Yet again, the last decoder layer of Gemma-2-9B (L39) experiences a drop. This figure further displays the pattern of middle layers sharing the most similarity, especially compared to early and late layers.
\begin{figure*}[h]
  \centering
  \includegraphics[width=0.7\textwidth]{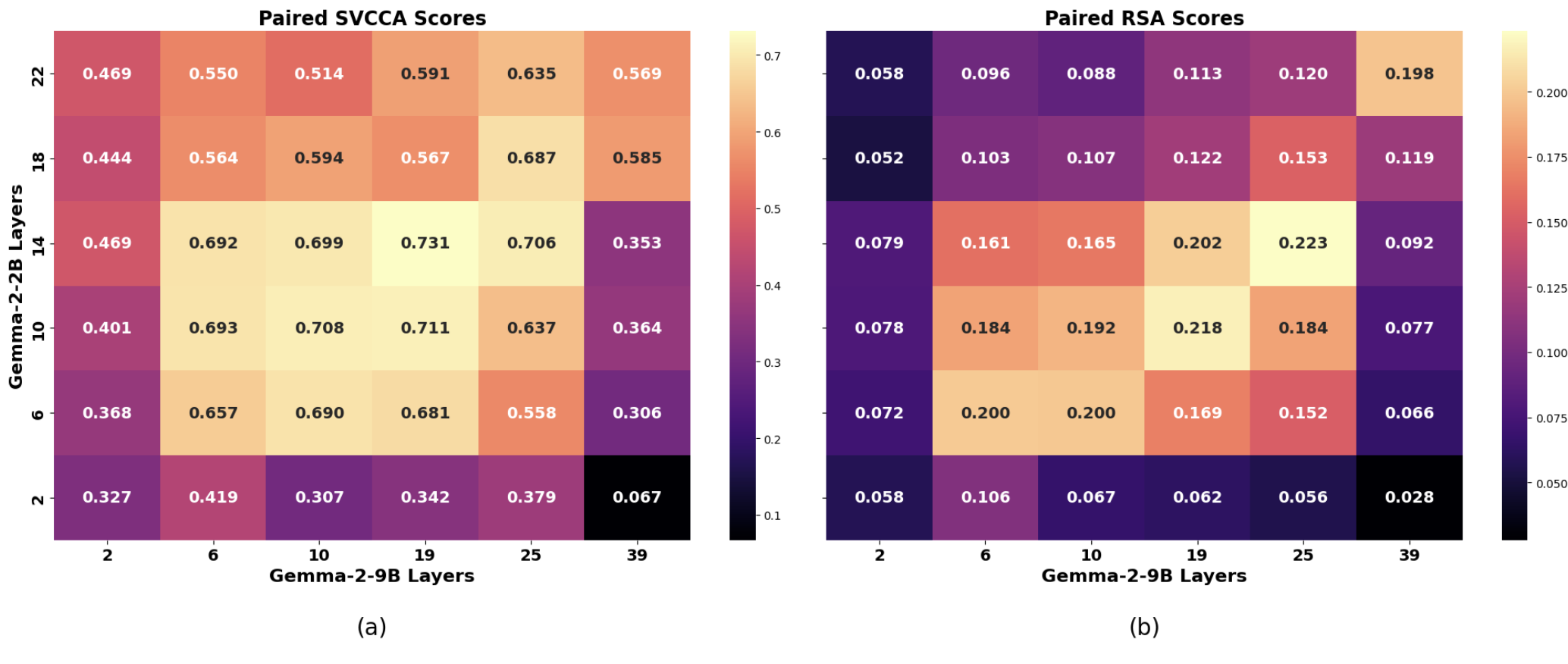} 
  \caption{(a) SVCCA and (b) RSA 1-to-1 paired scores of SAEs for layers in Gemma-2-2B vs Gemma-2-9B. Note the pattern of higher scores between the middle layer pairings indicating similarity in middle layers between both models.}
  \label{fig:example}
\end{figure*}

\textbf{Many-to-1 (Single run):} When we allow duplicates (Fig. 4a) the peak SVCCA softens to 0.69 (Gemma-2-2B L14 and Gemma-2-9B L19) and the mid-layer plateau narrows (0.54 – 0.66). RSA follows suit, topping out at around 0.18 - 0.2. This confirms that when features are matched more than once, the alignment scores drop slightly, but the overall pattern is maintained.
\begin{figure*}[h]
  \centering
  \includegraphics[width=0.7\textwidth]{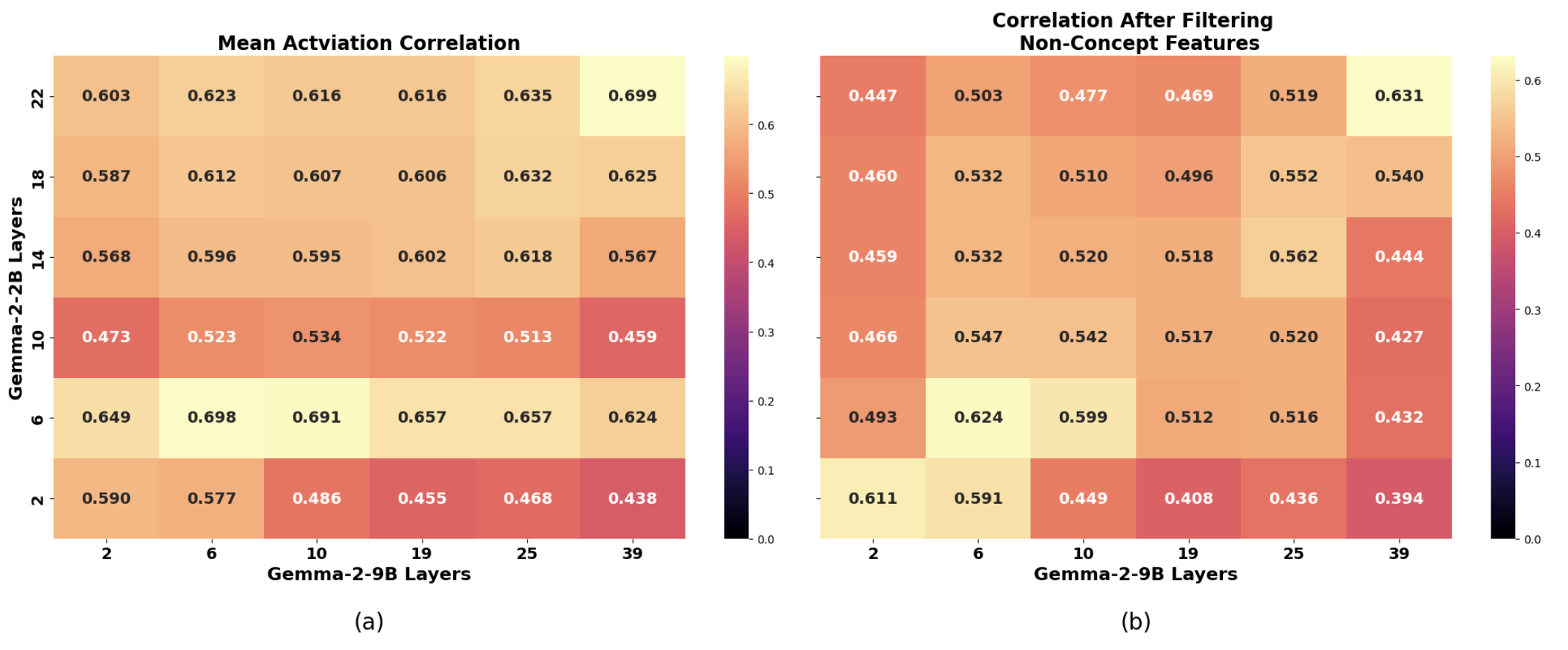} 
  \caption{(a) 1-to-1 Mean Activation Correlation before and (b) after filtering non-concept features for Gemma-2-2B vs Gemma-2-9B. Note these patterns generally contrast from those of the SVCCA and
RSA scores in Figure 2, indicating that these metrics each reveal different patterns not shown previously such as the pattern of middle layers between both models exhibiting higher correlation.}
  \label{fig:example}
\end{figure*}

\textbf{Many-to-1 (5 run mean):} Averaging five random initializations barely changes the picture (Fig. 6a): peak SVCCA = 0.69, peak RSA = 0.20. The variance across runs is $<$0.02 for every cell, indicating that the many-to-one procedure is stable, but still consistently lower than the ceiling of the 1-to-1 strategy.\\
\\
\textbf{Random-pair baselines and significance.}
Across the three experiments, the mean random SVCCA spans 0.005 – 0.034, with a majority of cells below 0.02 (Fig. 8a, Fig. 9a, Fig. 10a). Consequently, every empirical SVCCA score beyond the first residual layer lands in the 0.0 p-value bucket ($\leq$ 0.1\% chance of getting such a good alignment by random) (Fig. 8b, Fig. 9b, Fig. 10b). In other words, even the weakest observed alignment (SVCCA $\approxeq$ 0.30) is simply too strong to be by chance.\\
\\
\textbf{Effect of filtering non-concept features.}
Mean activation correlation before filtering peaks at 0.70 (L22 \& L39) and averages 0.60 ± 0.07 on the mid-layer block (Fig. 3a). After removing unimportant features (Fig. 3b) the pattern of middle layers having an increased correlation between the two models becomes more evident, while more surrounding random matches fall by 0.10 – 0.15, raising the peak correlation from 0.70 to 0.74. Another result to note is that early layers and late layers of both models share strikingly higher similarity compared to other layers; for example L2 of both models, L6 of both models, and Gemma-2-2B L22 \& Gemma-2-9B L39 all have the highest correlation (Fig. 3b, Fig. 5b, Fig. 7b). In other words, removing low-level features (such as punctuation) made strong alignments clearer and more meaningful, without just artificially boosting scores.

\begin{table*}[h]
\centering
\small
\begin{tabular}{lccc}
\toprule
\textbf{Overlapping Concept} & \textbf{Paired SVCCA Mean} & \textbf{Random Shuffling Mean} & \textbf{p-value} \\
\midrule
Emotion and Time & 0.62 & 0.13 &0.0  \\
Nature and People & 0.63 & 0.17 &0.0  \\
\bottomrule
\end{tabular}
\caption{Comparison of paired SVCCA, random shuffling mean, and p-values for reasonable pairs of concepts at layers 10 of both Gemma-2-2B and Gemma-2-9B.}
\label{tab:concepts good pairs}
\end{table*}
\begin{table*}[h]
\small
\centering
\begin{tabular}{lccc}
\toprule
\textbf{Overlapping Concept} & \textbf{Paired SVCCA Mean} & \textbf{Random Shuffling Mean} & \textbf{p-value} \\
\midrule
Country and People & 0.03 & 0.13 &0.02  \\
\bottomrule
\end{tabular}
\caption{Comparison of paired SVCCA, random shuffling mean, and p-values for bad pairs of concepts at layers 10 of Gemma-2-2B and layer 19 of Gemma-2-9B. These results indicate that these pairs are not encoded similarly}
\label{tab:concepts bad pair}
\end{table*}
\begin{table*}[h]
\small
\centering
\begin{tabular}{lccc}
\toprule
\textbf{Multi-token Concept} & \textbf{Paired SVCCA Mean} & \textbf{Random Shuffling Mean} & \textbf{p-value} \\
\midrule
Emotion and Time (L6 vs L2)& 0.27 & 0.02 &0.0  \\
Emotion and Time (L6 vs L10)& 0.53 & 0.02 &0.0 \\
\bottomrule
\end{tabular}
\caption{Comparison of paired SVCCA, random shuffling mean, and p-values for bad pairs of concepts at layers 6 of Gemma-2-2B and layer 2 and 10 of Gemma-2-9B. These results indicate that multi-token inputs are  encoded similarly.}
\label{tab:concepts multitokens}
\end{table*}
\subsection{Semantic Subspace Analysis}
\textbf{Semantic subspace alignment.}  
When we fix a single Gemma-2-2B layer and correlate every semantic-concept row against layers 2, 6, 10, 19, 25, and 39 for Gemma-2-9B, the same mid-on-mid pattern re-emerges: mid-stack layers in both models align best. Among the 2-2B sources we tried, the layer centered around L14 most consistently exhibited the highest SVCCA and RSA scores, reinforcing the idea that internal concept geometry more commonly converges in the middle of the networks. Full heat-maps for every source layer and metric are collected in Appendix B.

\textbf{Overlapping Semantic Subspaces}. When two semantic subspaces are combined, there are two different trends in the results based on the compatibility of the subspaces. For instance, combining the subspaces ``country'' and ``people'' in Table \ref{tab:concepts bad pair} which is a non-ideal pair results in low average SVCCA and RSA scores across the layers, indicating that there is insignificant correlation in how the models internally represent this subspace. This phenomenon could be caused by the fact that the models are unlikely to group these subspaces together during training. However, when the pair of subspaces being combined makes sense, such as ``nature'' and ``people'' in Table \ref{tab:concepts good pairs}, both the SVCCA and RSA scores are high, leading to the conclusion that both models represent these subspaces very similarly. All of the results are in Appendix \ref{appendix c}.

\textbf{Multi-token Semantic Subspaces}. Despite resource constraints limiting our evaluations to the "emotions time" subspace, our preliminary results on multi-token subspace (see Table \ref{tab:concepts multitokens}) provide key insights. Notably, high SVCCA scores remained in the early and middle layers, providing strong empirical evidence that models sometimes encode multi-token concepts. Furthermore, the SVCCA scores are drastically higher than "emotions" or "time" alone in the early layers (see Appendix B), indicating that earlier layers represent multi-token subspaces rather than single-token ones. This result challenges the popular, underlying assumption that models internally encode single-token concepts \citep{dehouck2023challenging, valois2024frame}. Hence, we believe that meaning is sometimes distributed across multiple tokens, and that semantic subspaces are the better level of analysis.  

\textbf{Distance metrics}. During layer-to-layer SAE feature analysis, we have used the Pearson correlation as done in \citet{lan2024sparse}. We tested other metrics such as the cosine similarity and euclidean distance; however, changing the similarity metric did not yield any statistically significant changes in the results, implying that the distance metric used does not affect the accuracy of our results. 

\section{Related Works}
\textbf{Superposition and Sparse Autoencoders.} Previous studies have shown that, when there are more features to be represented than available parameters, feature representations are distributed across multiple parameters, leading to polysemantic neurons \citep{elhage2022superposition}. Polysemanticity causes challenges in interpreting models, which is crucial for AI safety in identifying goal misgeneralization \citep{shah_goal_2022, langosco_goal_2022} as well as deceptive misalignment \citep{hubinger_sleeper_2024, greenblatt_alignment_2024}. For these reasons, Sparse Autoencoders (SAEs) have been used to transform polysemantic neuron activations into monosemantic feature neurons that usually correspond to one feature \citep{Makhzani2013kSparseA, cunningham2023sparseautoencodershighlyinterpretable, gao2024scalingevaluatingsparseautoencoders, rajamanoharan2024improvingdictionarylearninggated, rajamanoharan2024jumpingaheadimprovingreconstruction}. It is much easier to conduct quantitative interpretability studies on these monosemantic features. 
\\
\textbf{Feature Universality.} The existance of ``universal'' neurons across LLMs were first uncovered in a study of GPT-2 \citep{gurnee2024universalneuronsgpt2language}. Furthermore, previous studies that have performed quantitative analysis using SAEs to test for feature universality \citep{lan2024sparse, bricken2023monosemanticity} have shown universality in analogous features and representational features \citep{olah2020zoom, Yosinski2014, gurnee2024universalneuronsgpt2language, kornblith2019similarityneuralnetworkrepresentations}. These are not measures of ``true features'', which are stricter ground-truth features \citep{bricken2023monosemanticity} that represent atomic linear directions \citep{Till2023}.

Previous research on SAEs to test for feature universality \citep{lan2024sparse} has demonstrated that, after aligning neurons via mean activation correlation, there exists statistically significant alignment ($p < 0.05$) for almost all non-input layers. The middle layers exhibited the strongest correspondence, indicating that distinct LLMs learn a shared set of features. Beyond layer-wise comparisons, semantically defined subspaces were tested by filtering features whose top activating tokens match curated keyword lists linked to a conceptual category such as ``Emotions''or ``Time``. These subspaces yielded high SVCCA scores with $p\ll0.05$, illustrating that semantic concept feature groups are more consistent across models. 
\\
\textbf{Mechanistic Interpretability.} Interpreting neurons and MLP analysis have become increasingly popular \citep{foote2023neurongraphinterpretinglanguage, garde2023deepdecipheraccessinginvestigatingneuron} techniques to understand the inner workings of LLMs. Other mechanistic interpretability methods have used SAEs \citep{lan2024sparse, nanda2024decompsae} because of their more interpretable feature spaces.
\\

\section{Conclusion}
In this paper, we provided a comprehensive study on feature universality in models of different sizes. Our experiments reveal that, in many cases, internal representations are similar across models regardless of size. In addition, our semantic subspace experiments revealed that different models encode pairs of subspaces and multi-token subspaces similarly, further enforcing the concept of LLM universality. These findings are a key step in forming a comprehensive understanding of how LLMs internally function. \\  

\section{Limitations}
This study focused only on the Gemma-2-2B and Gemma-2-9B models due to resource constraints, particularly GPU availability and compute time. While this size comparison captures meaningful differences in model scale, further work could extend our approach to a wider range of architectures, parameter counts, tokenizers and layer comparisons. It is important to note that using models with the same tokenizer leads to higher quality pairings and thus higher accuracy results \citep{lan2024sparse}. Additionally, the number of SAE-derived subspaces we analyzed was limited to keep manual inspection and downstream evaluations manageable. Expanding this analysis to a larger, more diverse set of subspaces could help further characterize the extent and nature of feature space universality. Furthermore, our experiments with multi-token subspaces only tested a single subspace due to further resource limitations.

\bibliography{custom}

\clearpage
\theendnotes

\appendix
\label{sec:appendix}
\section{Additional Results}
\begin{figure}[H]
  \centering
  \includegraphics[width=0.475\textwidth]{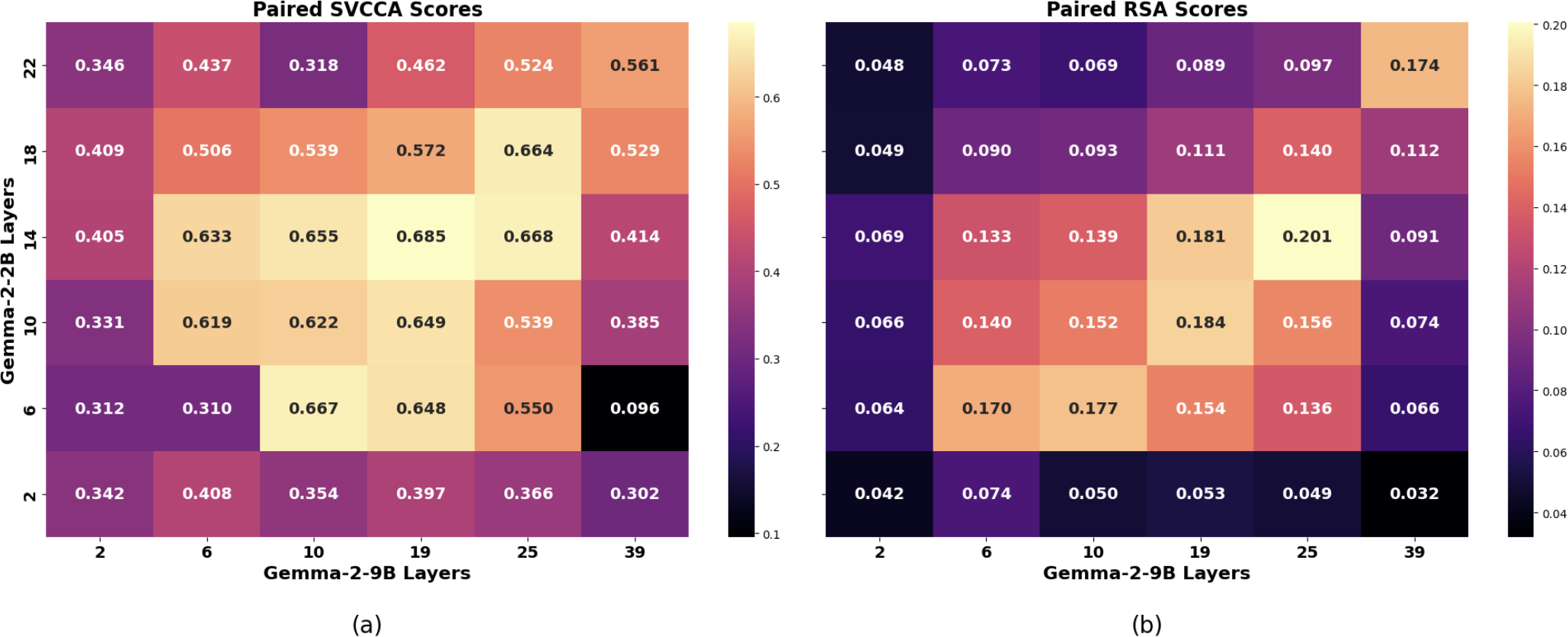} 
  \caption{(a) SVCCA and (b) RSA Many-to-1 paired scores of SAEs for layers in Gemma-2-2B vs Gemma-2-9B. Note the pattern of higher scores in the middle layers indicating similarity in middle layers between both models.}
  \label{fig:example}
\end{figure}
\begin{figure}[H]
  \centering
  \includegraphics[width=0.475\textwidth]{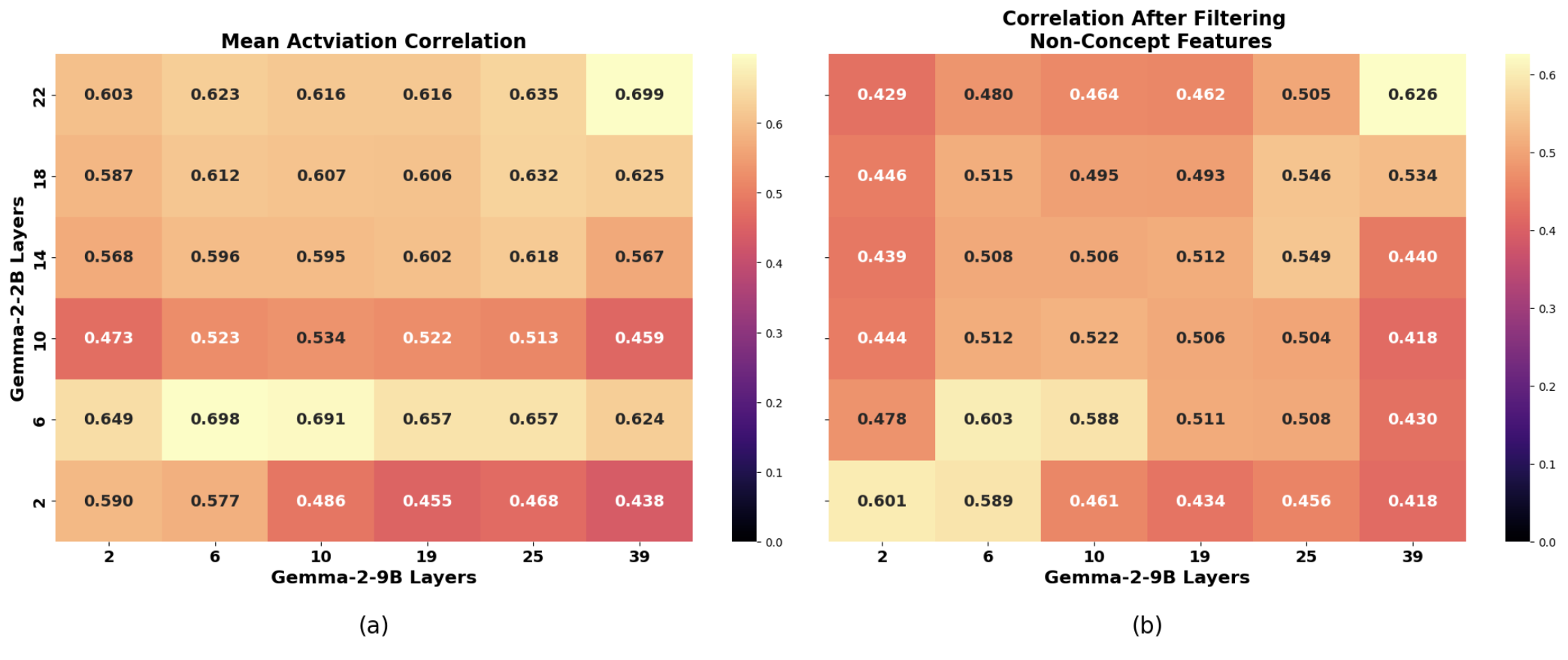} 
  \caption{(a) Many-to-1 Mean Activation Correlation before and (b) after filtering non-concept features for Gemma-2-2B vs Gemma-2-9B. Note these patterns generally contrast from those of the SVCCA and
RSA scores in Figure 2, indicating that these metrics each reveal different patterns not shown
previously.}
  \label{fig:example}
\end{figure}
\begin{figure}[H]
  \centering
  \includegraphics[width=0.475\textwidth]{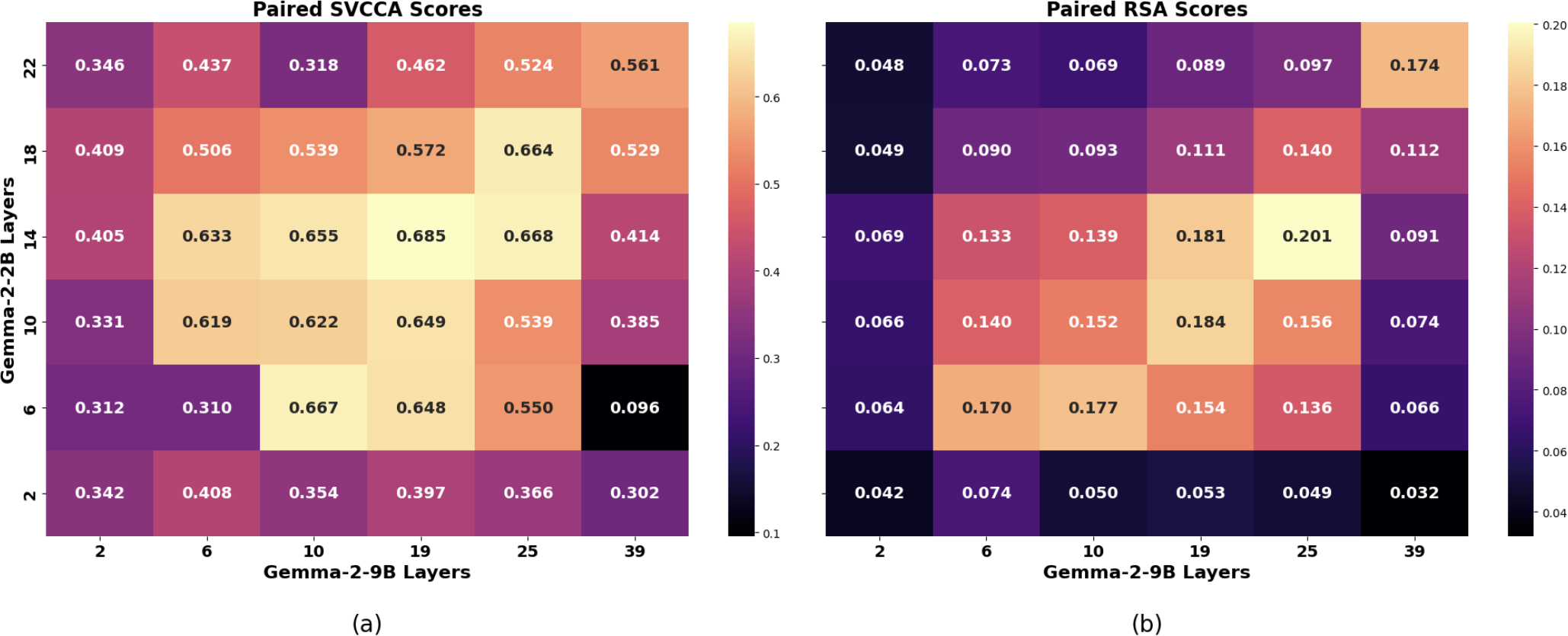} 
  \caption{(a) SVCCA and (b) RSA Many-to-1 (5 run average) paired scores of SAEs for layers in Gemma-2-2B vs Gemma-2-9B. Note the pattern of higher scores in the middle layers indicating similarity in middle layers between both models. This pattern is a trend in layer similarity between the three experiments as seen in Fig. 2a, Fig. 2b, Fig. 10a, and Fig. 4b.}
  \label{fig:example}
\end{figure}
\begin{figure}[H]
  \centering
  \includegraphics[width=0.475\textwidth]{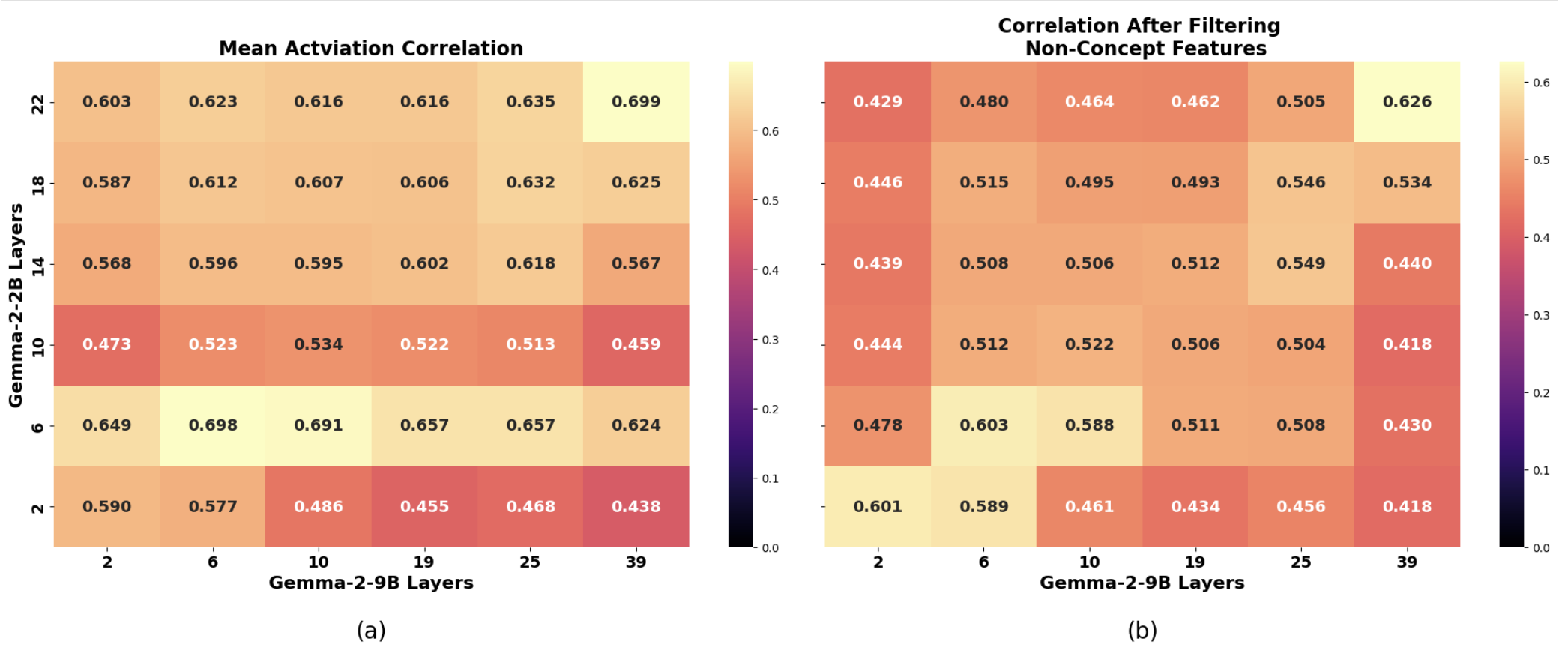} 
  \caption{(a) Many-to-1 (5 run average) Mean Activation Correlation before and (b) after filtering non-concept features for Gemma-2-2B vs Gemma-2-9B. Note these patterns generally contrast from those of the SVCCA and
RSA scores in Figures 2, 4, and 6, indicating that these metrics each reveal different patterns not shown
previously.}
  \label{fig:example}
\end{figure}
\begin{figure}[H]
  \centering
  \includegraphics[width=0.475\textwidth]{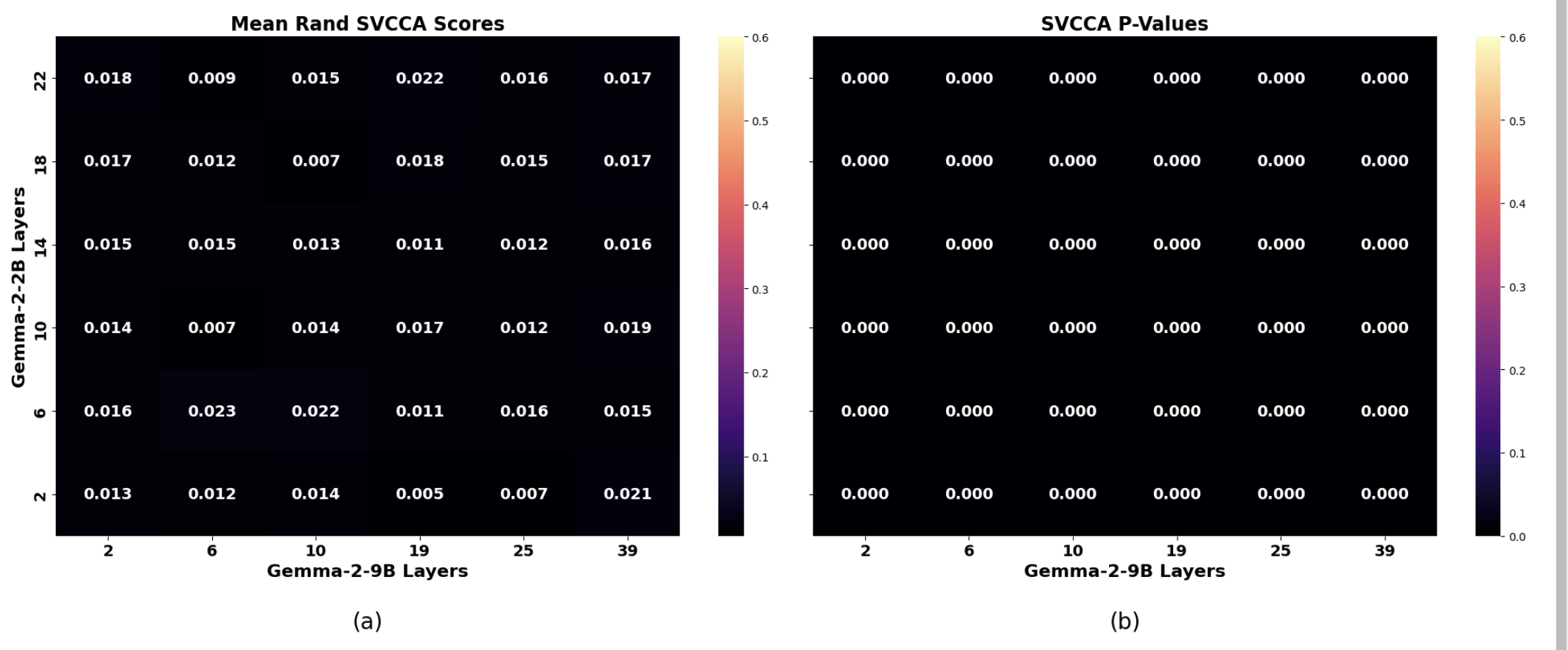} 
  \caption{(a) Mean Randomly Paired SVCCA 1-to-1 scores and (b) SVCCA 1-to-1 P-values of SAEs for layers in Gemma-2-2B vs Gemma-2-9B.}
  \label{fig:example}
\end{figure}
\begin{figure}
  \centering
  \includegraphics[width=0.475\textwidth]{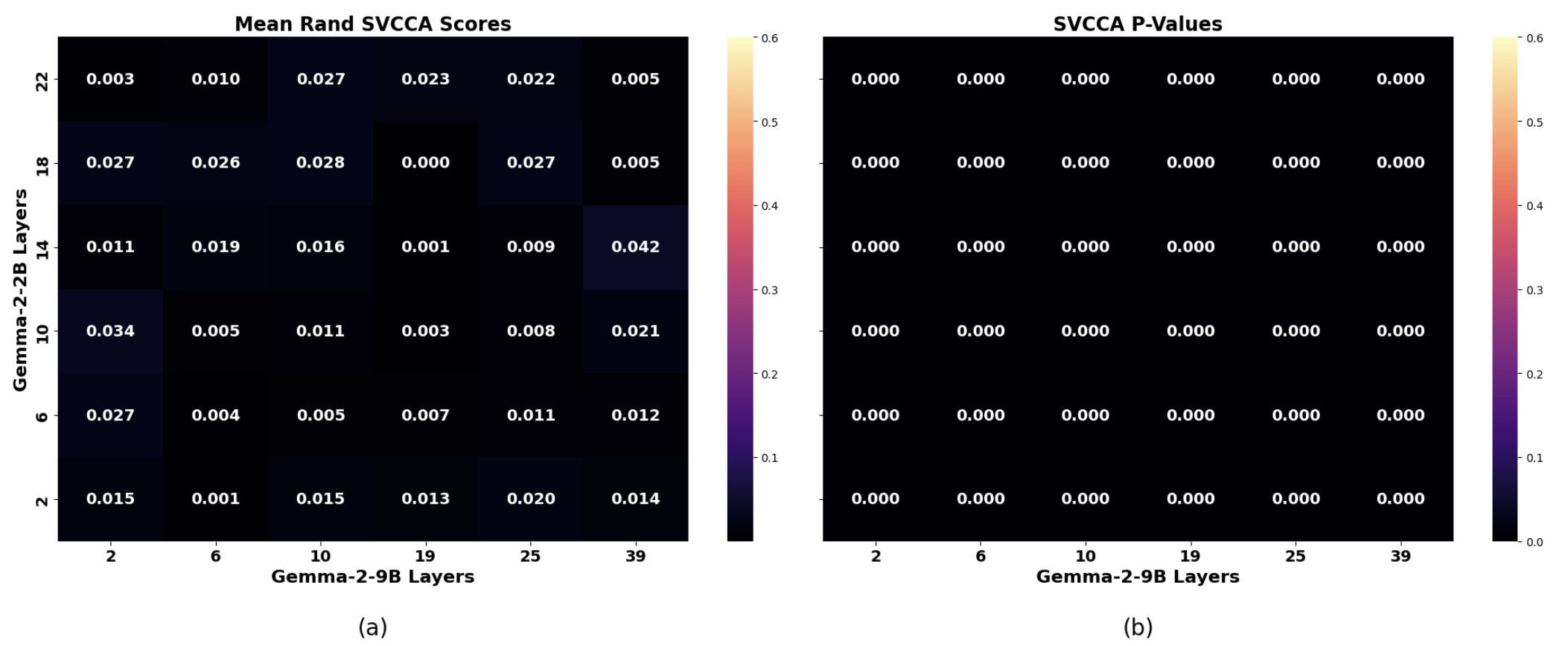} 
  \caption{(a) Mean Randomly Paired SVCCA Many-to-1 scores and (b) SVCCA Many-to-1 P-values of SAEs for layers in Gemma-2-2B vs Gemma-2-9B.}
  \label{fig:example}
\end{figure}
\begin{figure}[H]
  \centering
  \includegraphics[width=0.475\textwidth]{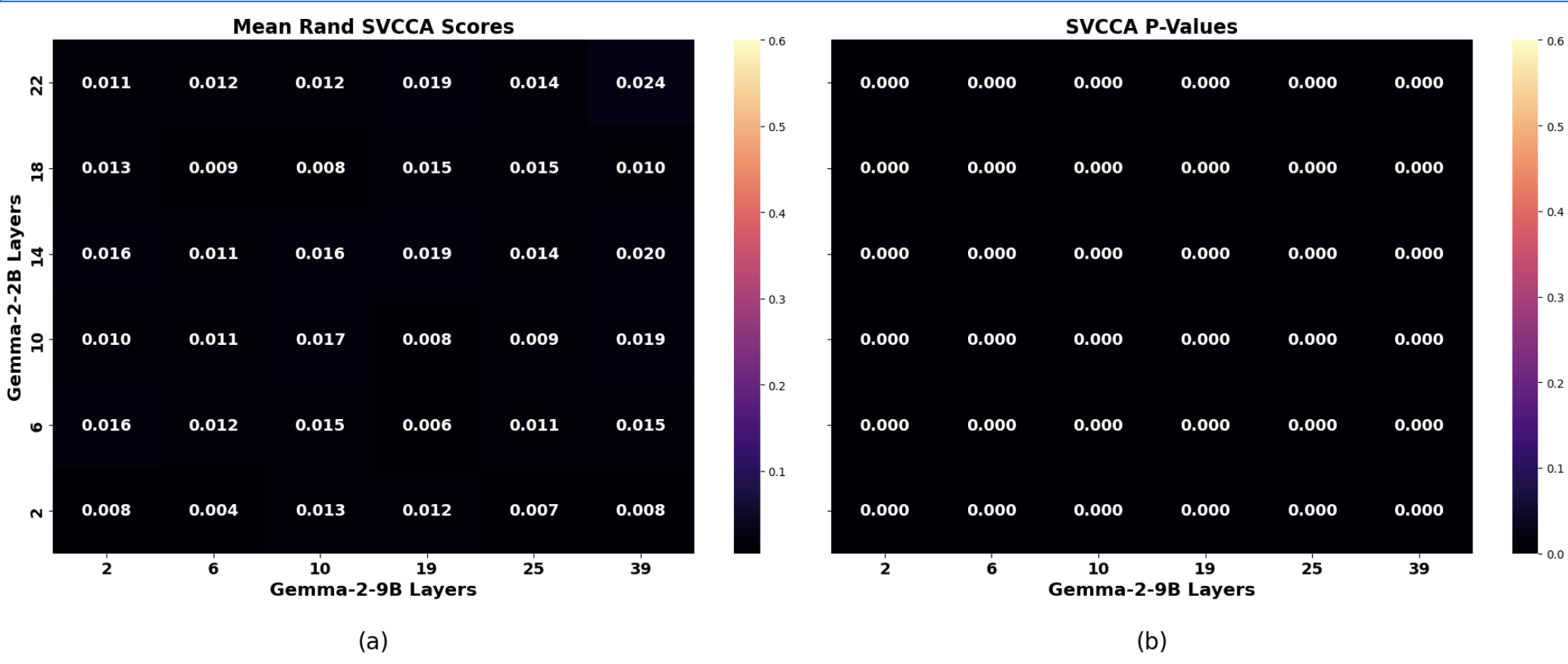} 
  \caption{(a) Mean Randomly Paired SVCCA Many-to-1 scores and (b) SVCCA Many-to-1 (5 run average) P-values of SAEs for layers in Gemma-2-2B vs Gemma-2-9B.}
  \label{fig:example}
\end{figure}

\section{Semantic Subspace Similarity (Gemma-2-2B $\rightarrow$ Gemma-2-9B).}
\label{appendix b}
    The following figures visualize concept-wise alignment from fixed layers of Gemma-2-2B to all layers of Gemma-2-9B. Each heatmap row corresponds to a semantic category (e.g., \textit{Emotions}, \textit{Biology}) and each column is a layer in Gemma-2-9B.

    We show both SVCCA and RSA 1-to-1 results for 2-2B source layers L6, L10, L14, and L17. These help assess which layers in Gemma-2-9B best match the concept geometry of the smaller model. For most layers, similarity scores peak in the mid-stack (L10–L19), further supporting the cross-model alignment trend observed in Section~4.

    \textit{* Some concepts may not appear across all rows if they lacked sufficient matched features or token coverage.}

  \vspace{1em}
\begin{figure}[H]
  \centering
  \includegraphics[width=0.475\textwidth]{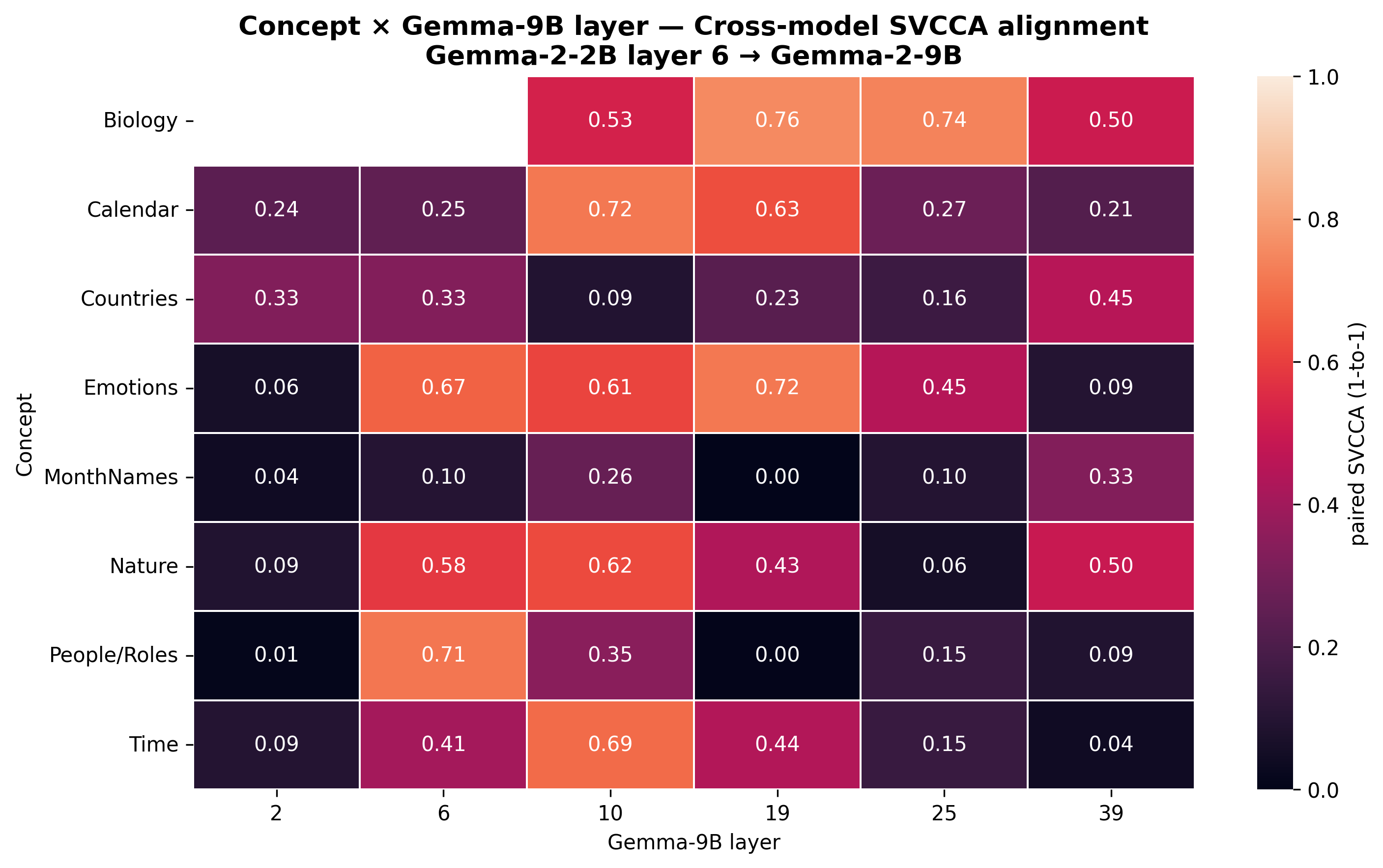}
  \caption{Scores rise into the mid-layers, peaking near 0.40 at L10–L19.}
  \label{fig:semantic_l6}
\end{figure}

\begin{figure}[H]
  \centering
  \includegraphics[width=0.475\textwidth]{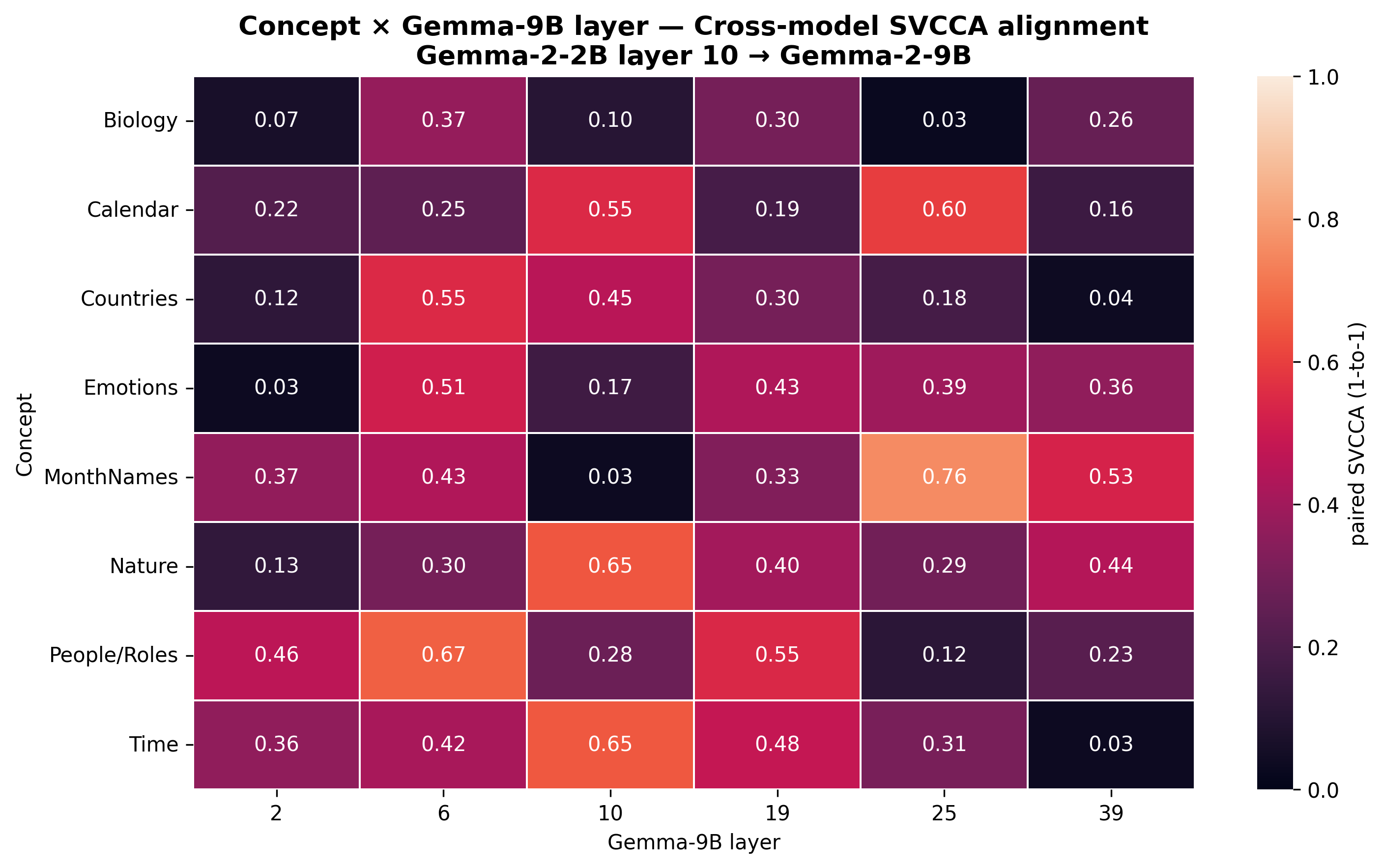} 
\caption{Mid-layer alignment strengthens; several concepts surpass 0.60.}
  \label{fig:example}
\end{figure}
\begin{figure}[H]
  \centering
  \includegraphics[width=0.475\textwidth]{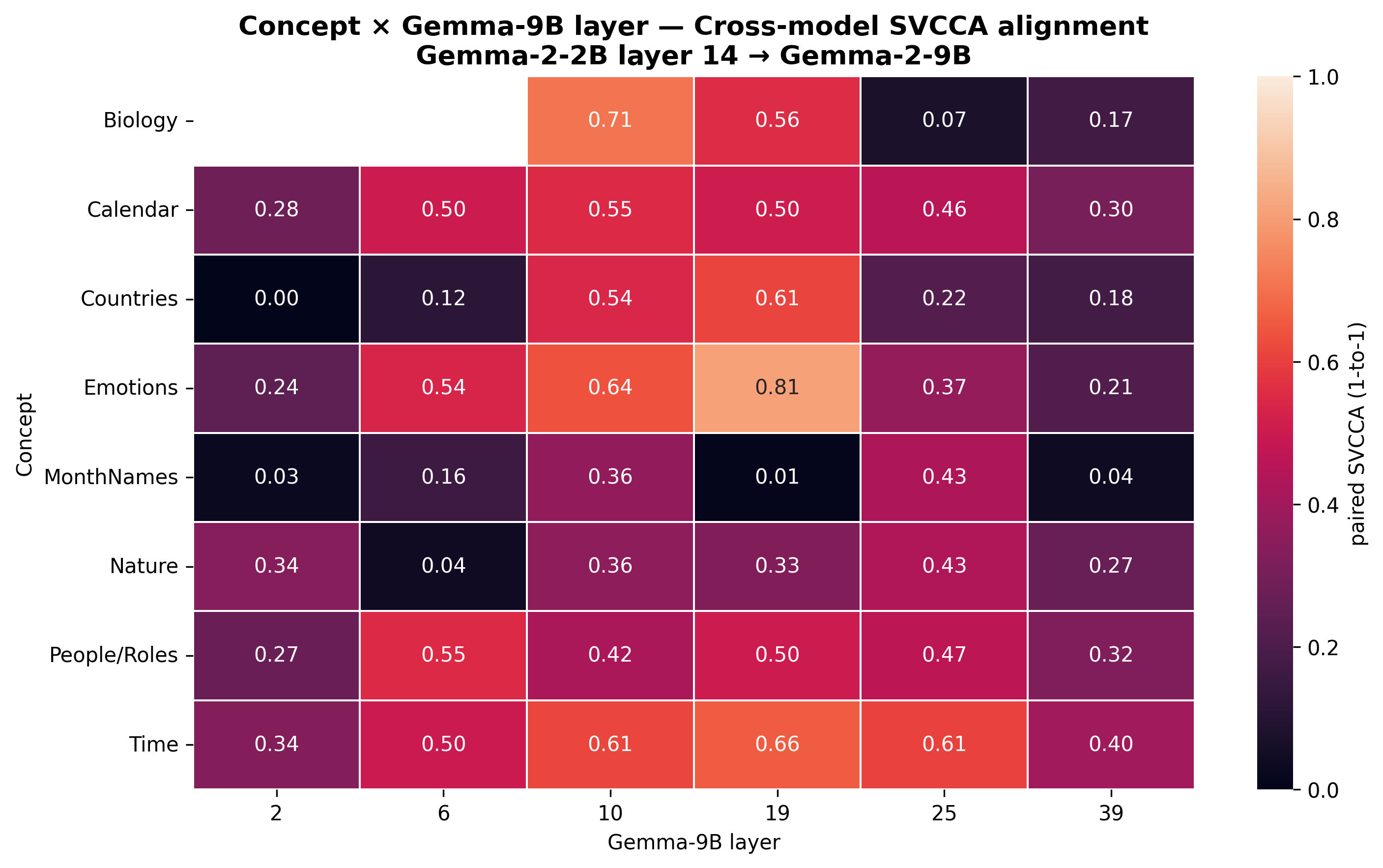} 
\caption{Highest similarity sits in the center stack; edge layers lag.}
  \label{fig:example}
\end{figure}
\begin{figure}[H]
  \centering
  \includegraphics[width=0.475\textwidth]{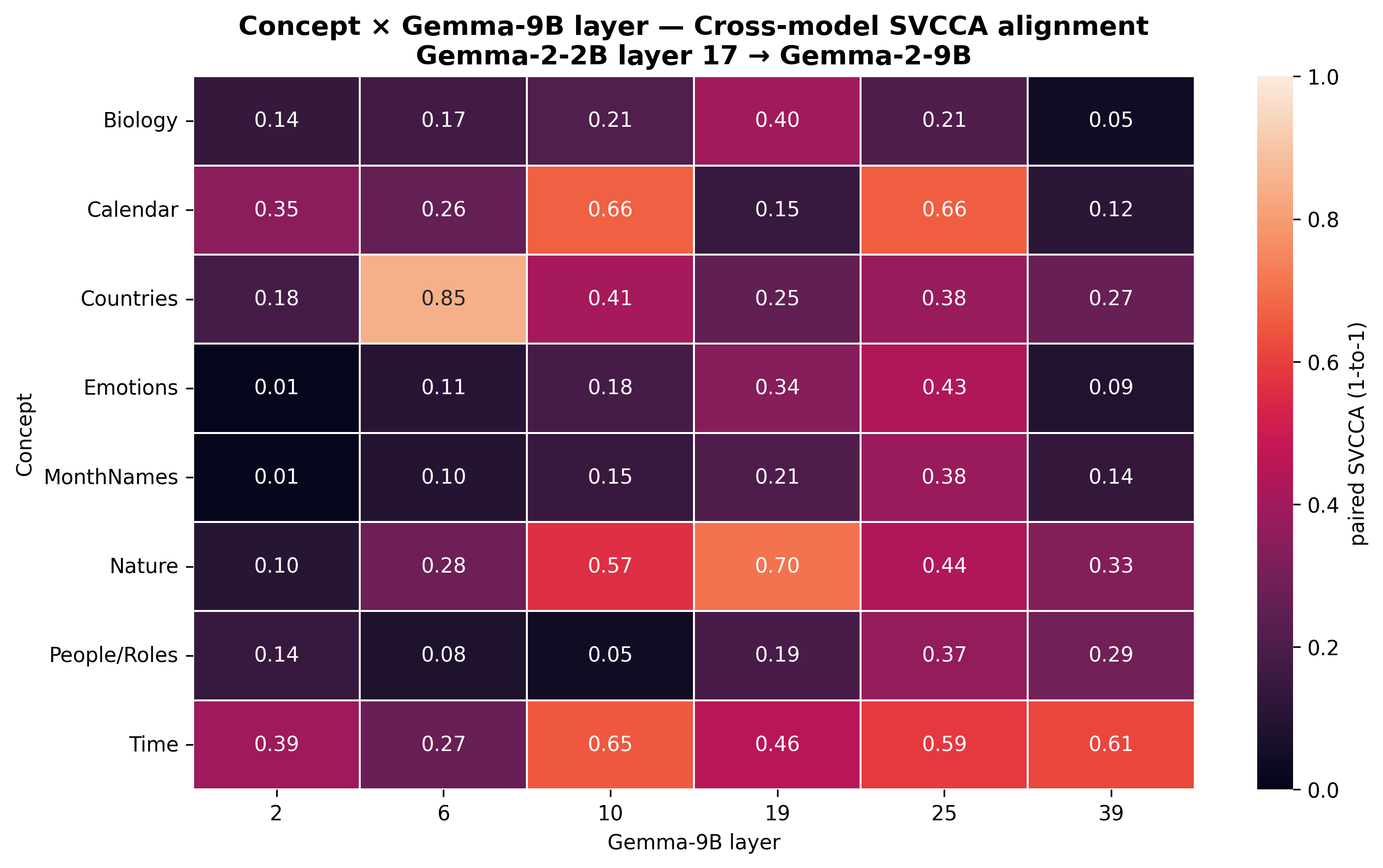} 
\caption{Alignment remains centered; deep and early layers score lower.}
  \label{fig:example}
\end{figure}
\begin{figure}[H]
  \centering
  \includegraphics[width=0.475\textwidth]{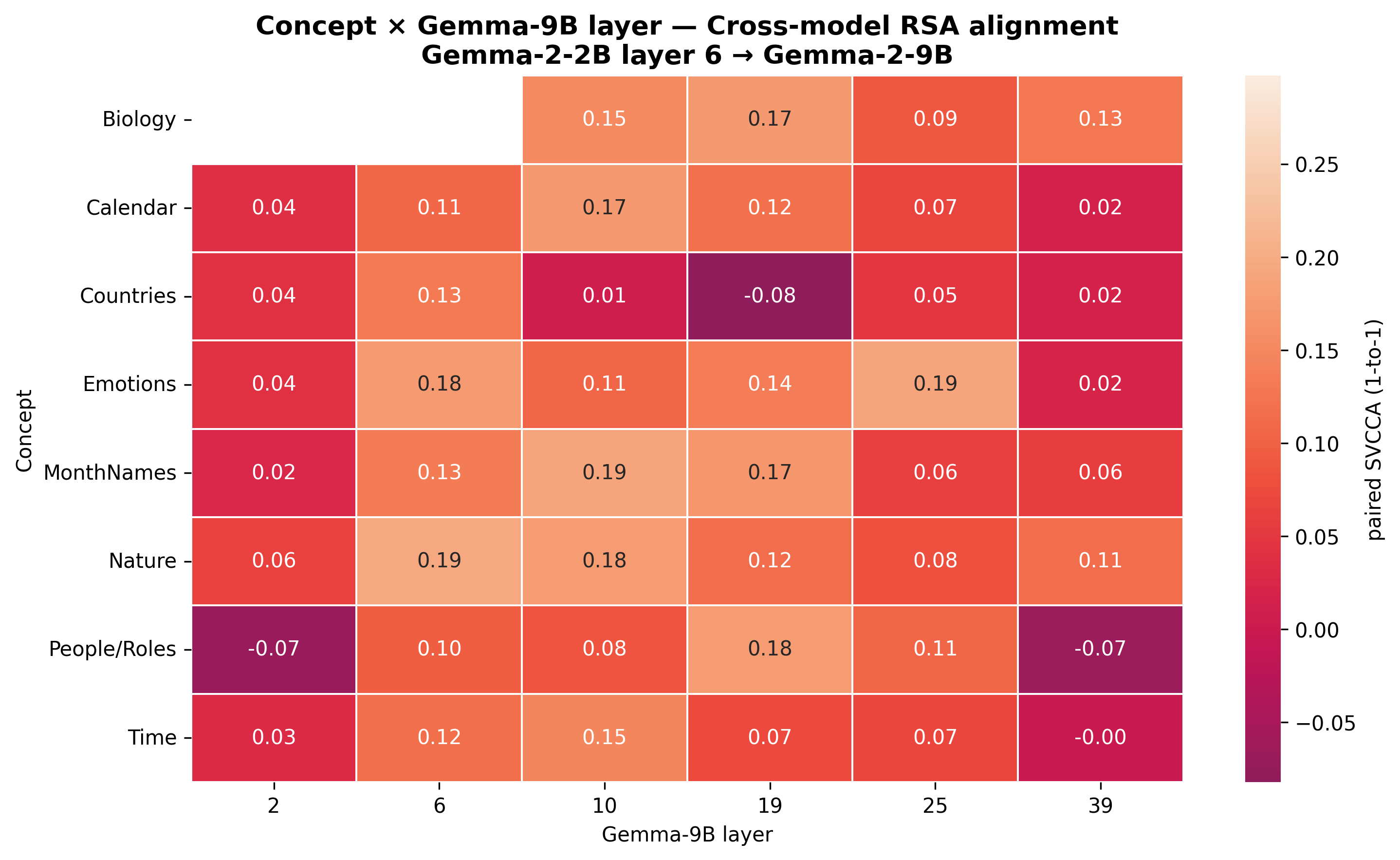} 
\caption{Mid-layer RSA peaks near 0.17; edge layers stay low. \textit{People/Roles} dips below 0 at L2.}
  \label{fig:example}
\end{figure}
\begin{figure}[H]
  \centering
  \includegraphics[width=0.475\textwidth]{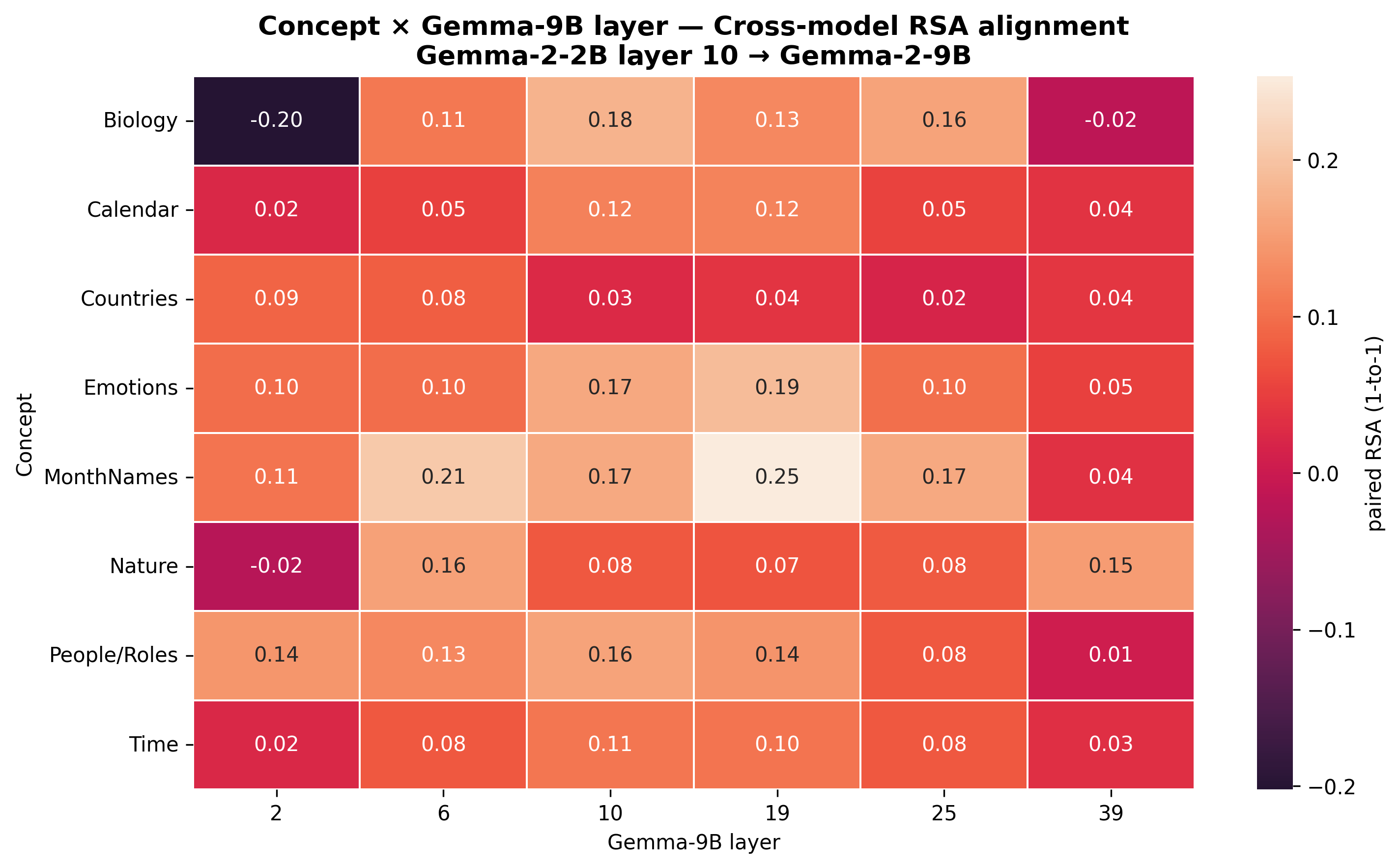} 
\caption{Mid-to-deep pairs score higher: \textit{Biology}, \textit{Month-Names} reach $\sim$0.25 at L19.}
  \label{fig:example}
\end{figure}
\begin{figure}[H]
  \centering
  \includegraphics[width=0.475\textwidth]{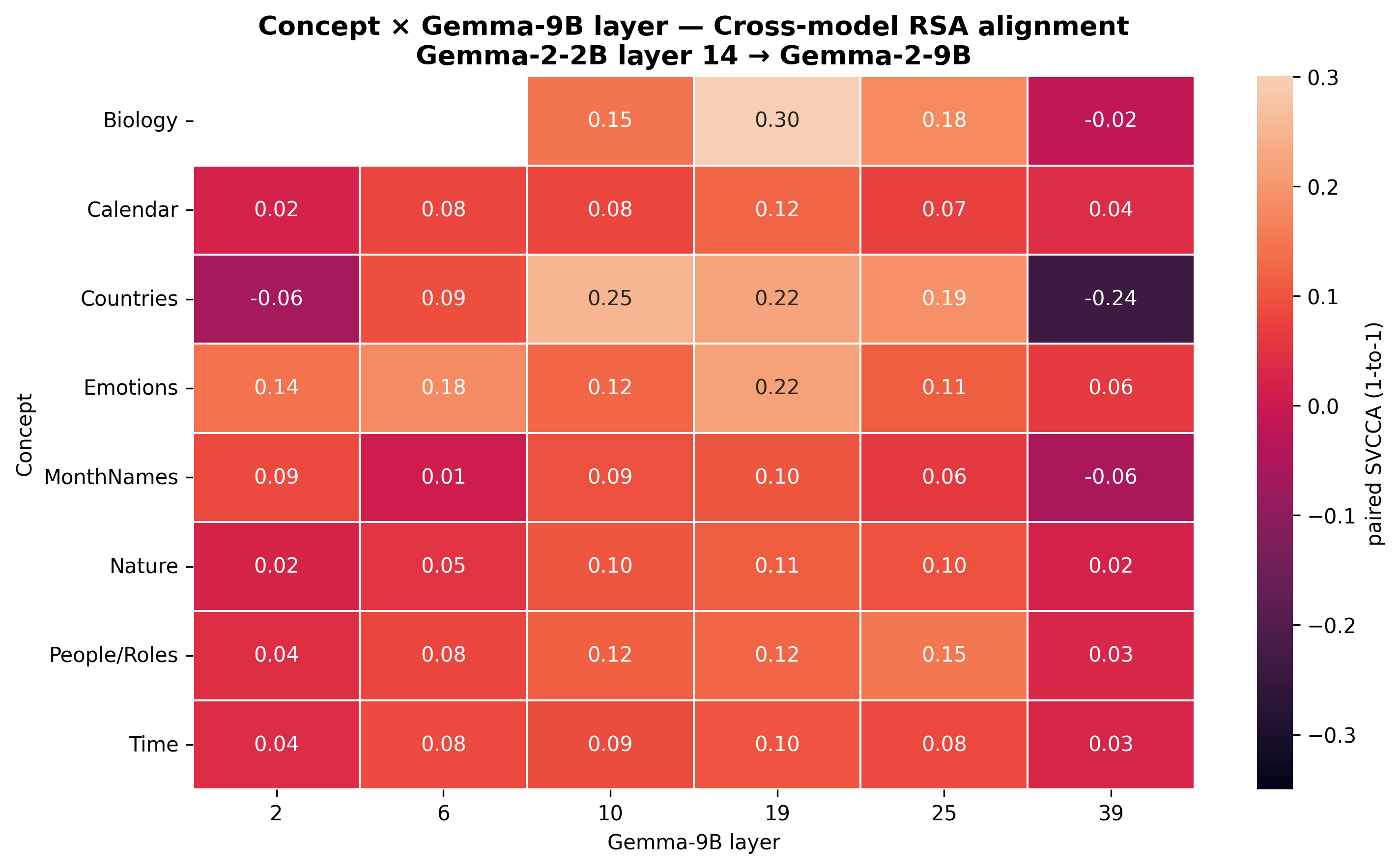} 
\caption{Highest scores at 9B L19: \textit{Biology} $\sim$0.30, \textit{Countries} $\sim$0.25. L2 and L39 remain near zero.}
  \label{fig:example}
\end{figure}
\begin{figure}[H]
  \centering
  \includegraphics[width=0.475\textwidth]{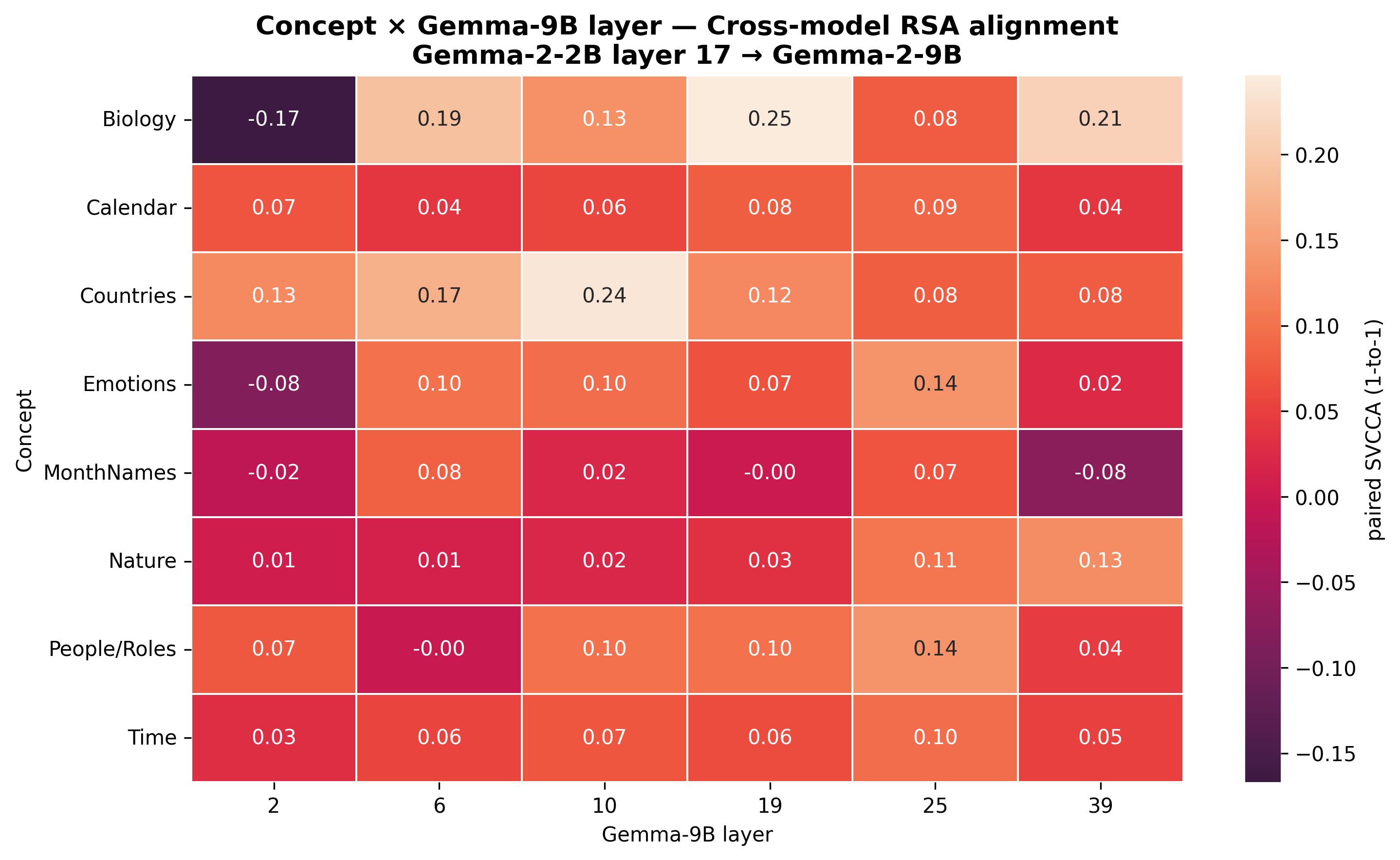} 
\caption{Strongest alignment at L19; outer layers weak or negative.}
  \label{fig:example}
\end{figure}

\section{SVCCA by concept in Gemma-2-2B and Gemma-2-9B.}
\label{appendix c}
\begin{figure}[H]
  \centering
  \includegraphics[width=0.475\textwidth]{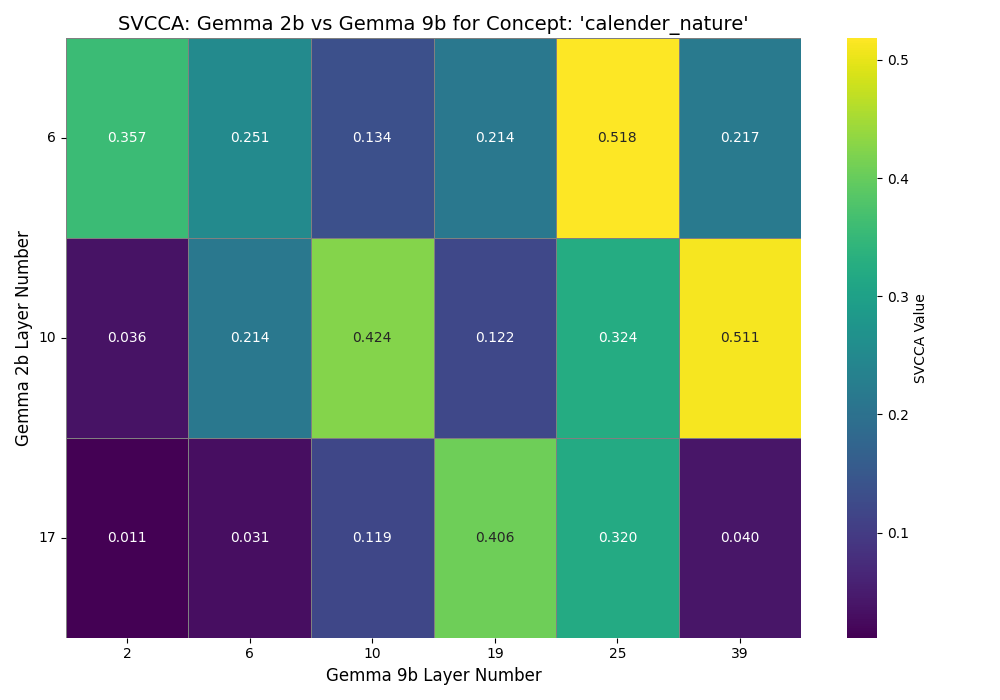} 
  \caption{paired SVCCA 1-to-1 for calendar-nature concept}
  \label{fig:example}
\end{figure}
\begin{figure}[H]
  \centering
  \includegraphics[width=0.475\textwidth]{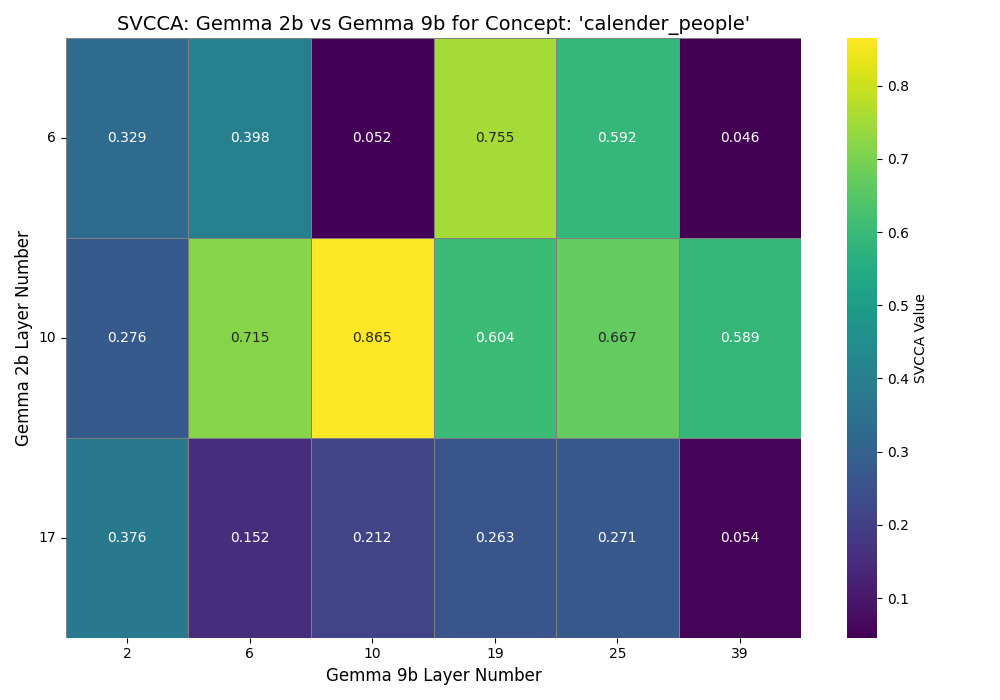} 
  \caption{paired SVCCA 1-to-1 for calendar-people concept}
  \label{fig:example}
\end{figure}
\begin{figure}[H]
  \centering
  \includegraphics[width=0.475\textwidth]{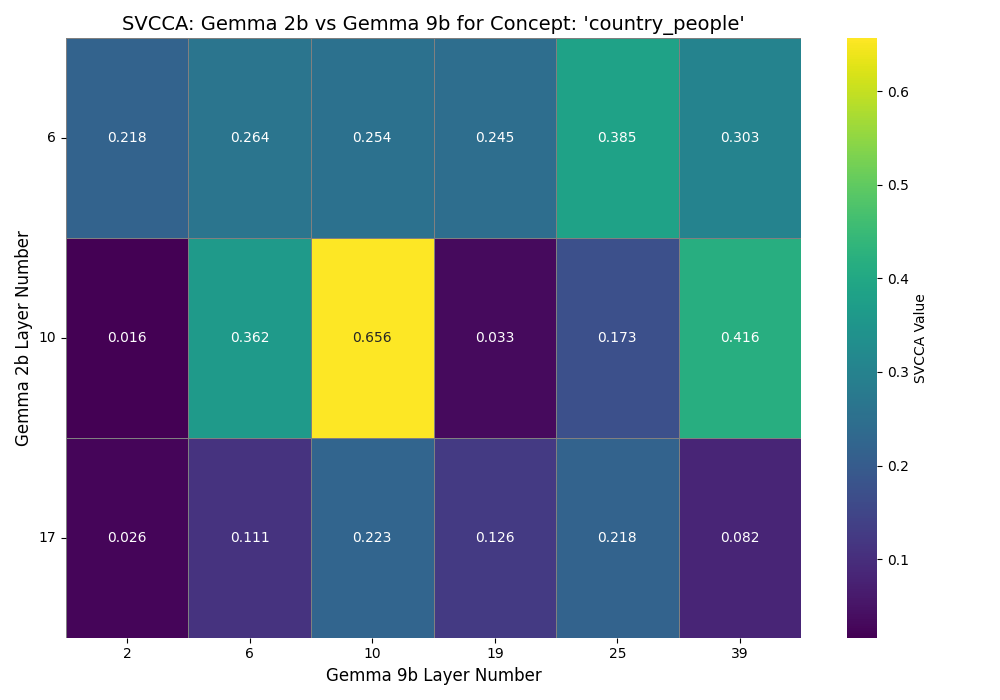} 
  \caption{paired SVCCA 1-to-1 for country-people concept}
  \label{fig:example}
\end{figure}
\begin{figure}[H]
  \centering
  \includegraphics[width=0.475\textwidth]{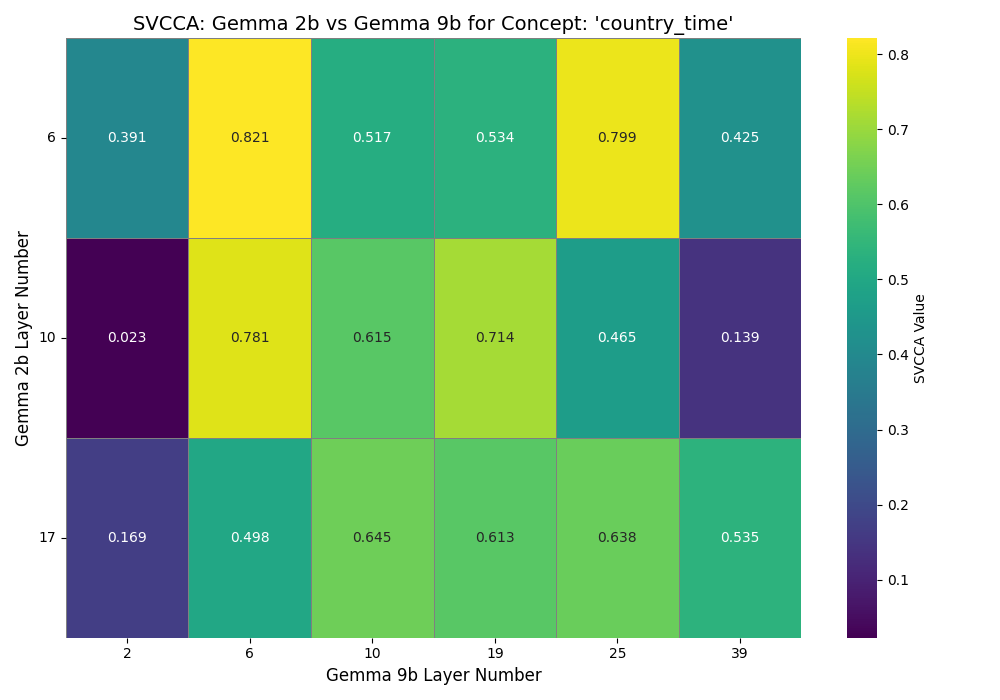} 
  \caption{paired SVCCA 1-to-1 for country-time concept}
  \label{fig:example}
\end{figure}
\begin{figure}[H]
  \centering
  \includegraphics[width=0.475\textwidth]{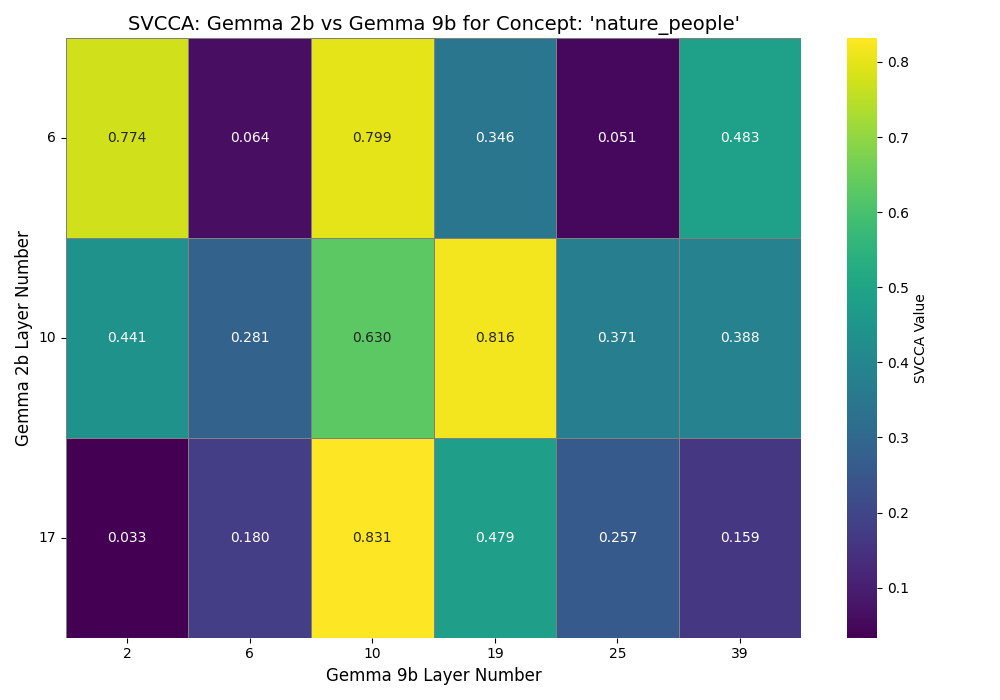} 
  \caption{paired SVCCA 1-to-1 for nature-people concept}
  \label{fig:example}
\end{figure}
\begin{figure}[H]
  \centering
  \includegraphics[width=0.475\textwidth]{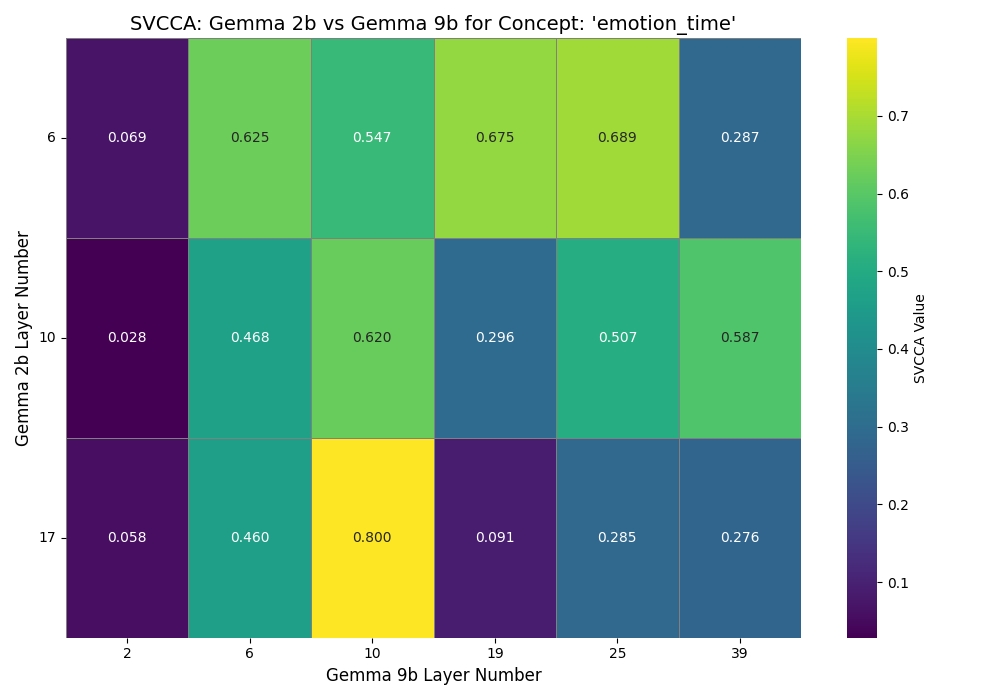} 
  \caption{paired SVCCA 1-to-1 for emotion-time concept}
  \label{fig:example}
\end{figure}

\end{document}